\documentclass{article}
\usepackage{spconf,amsmath,graphicx}

\usepackage[colorlinks,linkcolor=magenta,citecolor=cyan]{hyperref}
\usepackage{algorithm}
\usepackage{algpseudocode}
\usepackage{booktabs}
\usepackage{xcolor}

\title{Towards Robust Long-Context Understanding of Large Language Model via Active Recap Learning}
\name{Chenyu Hui$^1$\sthanks{Corresponding author.}}

\address{$^1$School of Microelectronics\\
Xi'an Jiaotong University\\
Xi'an, China\\
2687123206@stu.xjtu.edu.cn}
\begin{document}
%
\maketitle

\begin{abstract}
In this paper, we propose active recap learning (ARL), a framework for enhancing large language model (LLM) in understanding long contexts. ARL enables models to revisit and summarize earlier content through targeted sequence construction during continued pretraining and retrospective summarization at inference. First, we identify key tokens in prepared long context based on loss gaps between long and short forward contexts and find the most relevant preceding paragraphs, then summarize them using an LLM. Second, ARL equips models with the ability to autonomously generate and utilize these retrospective summaries during inference, thereby establishing a recursive memory mechanism across paragraphs. Experimental results show substantial gains, with ARL achieving an improvement of up to 129.7\% on RULER and 11--13\% on LongBench. Overall, ARL offers a simple yet effective continued pretraining-based approach to strengthen long-context understanding, advancing scalable memory augmentation in LLM. Code is available at \href{https://github.com/nanfangxiansheng/TOWARDS-ROBUST-LONG-CONTEXT-UNDERSTANDING-OF-LARGE-LANGUAGE-MODEL-VIA-ACTIVE-RECAP-LEARNING}{CODE}
\end{abstract}


\begin{figure*}[t]
	\centering
	\includegraphics[width=\linewidth]{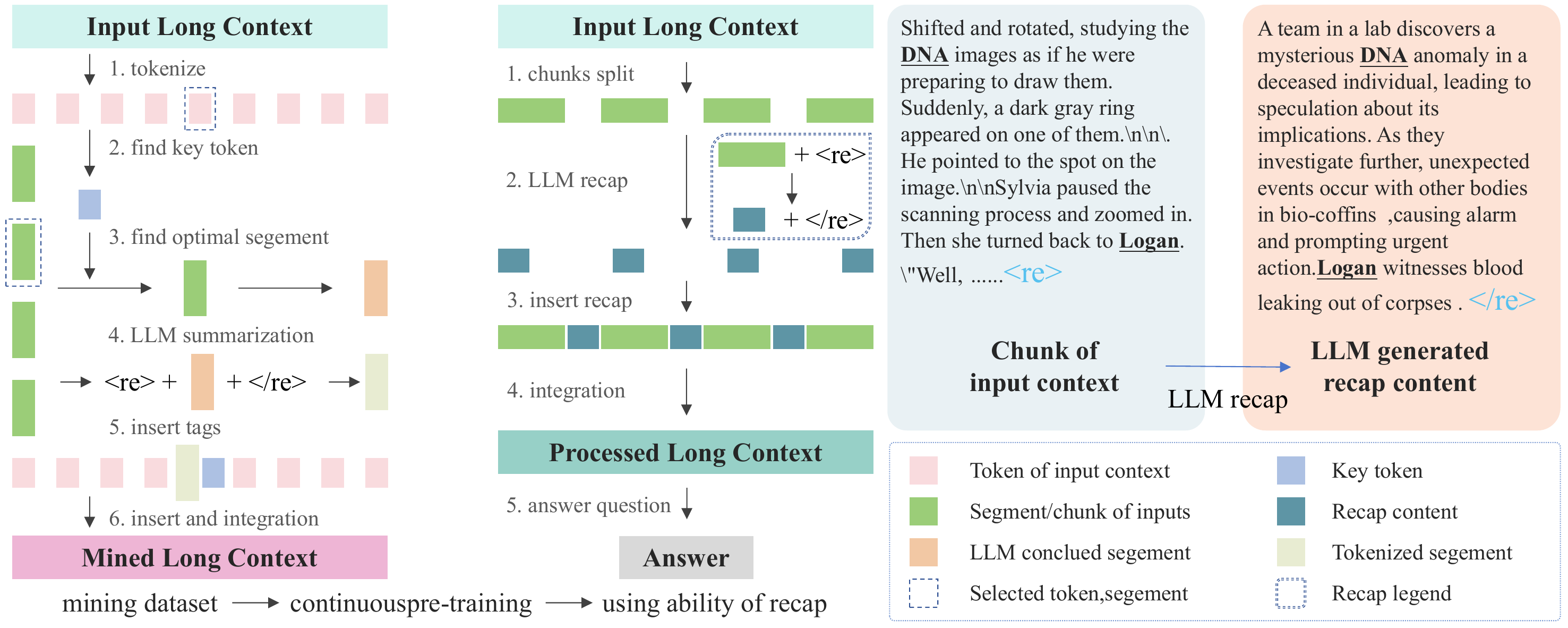}
	\caption{Overview framework of our proposed ARL with a chunk generating recap example.}
	\label{figure1}
\end{figure*}

\section{Introduction}
Long-context understanding is a core capability of large language model (LLM), reflected not only in their ability to process extended text sequences but also in their capacity to comprehend, reason over, and leverage complex information within them. Robust long-context understanding is therefore critical for applying LLM to real-world scenarios such as legal documents \cite{godbole-1}, academic manuscripts \cite{ma-2}, financial reports \cite{koval-3}, and large codebases \cite{li-4}. However, in practice, issues such as performance degradation under ultra-long inputs, memory constraints, and computational scalability often lead to suboptimal outcomes.

Current research on enhancing long-context understanding can be broadly categorized into three directions: length extrapolation, efficient attention, and context compression. The first line of work, which extends positional encoding or employs continued pretraining, often yields limited improvements while incurring high computational costs \cite{xiong-5}. The second line, which introduces linear or kernelized attention mechanisms, typically requires training models from scratch and encounters challenges in parallelization and scalability \cite{katharopoulos-6,zhao-6}. The third line, which relies on token compression or external memory modules, tends to disrupt the generation process and exhibits limited compatibility with mainstream frameworks \cite{wang-7}. Despite these advances, existing state-of-the-art methods have yet to achieve a balance of efficiency, generalization, and usability, leaving LLM performance on long-text tasks far from satisfactory.

To address these challenges, we propose active recap learning (ARL), a simple yet effective framework that strengthens a model’s ability to recall and exploit relevant context from earlier portions of long texts. As illustrated in Fig.~\ref{figure1}, ARL neither increases model capacity nor compresses information. Instead, it enables the model to perform active recapping during both continued pretraining and inference, thereby establishing a recursive memory mechanism. Concretely, we first construct novel pretraining sequences by identifying key tokens whose predictions are highly sensitive to the truncation of preceding context. For these tokens, we extract the most informative early segments and summarize them using an LLM. The resulting summaries are reinserted into the long text with special tags, allowing the model to learn context-recall patterns. During inference, the input text is segmented, and the same tagging mechanism prompts the model to generate a summary for each segment. These summaries are then propagated segment by segment, simulating a memory function that preserves coherence and consistency throughout long texts. The main contributions of this paper can be summarized as follows:
\begin{itemize}
	\item We introduce the long-short gap metric for high-specificity context tagging, enabling precise identification of critical information.
	\item We design a two-stage contextual extraction approach to identify optimal segments and refine LLM processing of long inputs.
	\item We develop an inference-time recap agent capable of generating local summaries and synthesizing global context, thereby maximizing the effectiveness of the recap mechanism.
\end{itemize}

\section{Proposed Method}
We propose active recap learning (ARL), a framework designed to improve long-range context modeling in large language models. ARL operates in three stages: (1) identifying key tokens using the long-short gap metric, (2) retrieving the most relevant earlier segments via loss-based retrieval, and (3) refining these segments into concise summaries. The augmented sequences are then used for continued pretraining, while a lightweight recap agent applies the same mechanism at inference to segment inputs and propagate summaries, enabling effective contextual recall across long texts.

\subsection{Recap Supervision}
To enhance long-context modeling, we first identify tokens that strongly depend on distant information. We introduce the long-short gap (LSG) metric, which measures a token’s reliance on long-range context by comparing prediction probabilities under two settings: (1) the full context and (2) a truncated short context. Formally, for token $x_i$:
\begin{equation}
	I_{\mathrm{LSG}}(x_i,\theta) = \frac{P_\theta(x_i \mid \boldsymbol{l}_i)}{P_\theta(x_i \mid \boldsymbol{s}_i)},
\end{equation}
where $P_\theta(x_i \mid \boldsymbol{l}_i)$ and $P_\theta(x_i \mid \boldsymbol{s}_i)$ denote the model’s predicted probabilities with long and short contexts, respectively. A higher score indicates stronger dependence on earlier content.We then select the Top-K tokens with the highest LSG scores as key tokens, which serve as anchors for retrieving  relevant segments from preceding text.

Given the key tokens identified by the LSG metric, we aim to locate the most informative preceding segment that enhances the prediction of each key token. Candidate segments are extracted from earlier parts of the document via a sliding window, aligned to sentence boundaries for coherence.

For a key token $x_i$, the optimal segment $c^*$ is defined as:
\begin{equation}
	c^* = argmax_{c \in \mathcal{C}} P_\theta \big(x_i \mid \text{Insert}(c, \mathcal{D}, i)\big),
\end{equation}
where $\text{Insert}(c, \mathcal{D}, i)$ denotes inserting segment $c$ before $x_i$ in document $\mathcal{D}$.

This selection ensures that the chosen context maximally improves the model’s confidence in predicting the target token, thereby reinforcing long-range dependency modeling. The primary procedures of finding the most relevant remote previous text fragment are shown in Algorithm~\ref{algorithm1}.

\begin{algorithm}
	\caption{: Mining Most Relevant Segments}
	\label{algorithm1}
	\begin{algorithmic}[1]
		\Require 
		\Statex $\text{text}$: Input document
		\Statex $\text{top\_k}$: Number of key tokens to select
		\Statex $\text{window\_size}$: Sliding window size (in characters)
		\Statex $\text{step\_size}$: Step size for window traversal
		\Ensure 
		\Statex A list of refined recap segments
		
		\State $\text{key\_tokens} \gets \text{SelectTopKTokens}(text, top\_k)$;
		\For{$x_i \in \text{key\_tokens}$}
		\State $best\_context \gets \text{None},\ best\_score \gets 0$;
		\For{$start = 0$ \textbf{to} $\text{len(text)}$ \textbf{step} $step\_size$}
		\State $c \gets \text{text}[start : start + window\_size]$;
		\State $text' \gets \text{Insert}(c, text, i)$;
		\State $p \gets P_\theta(x_i \mid text')$;
		\If{$p > best\_score$}
		\State $best\_context \gets c,\ best\_score \gets p$;
		\EndIf
		\EndFor
		\State $refined \gets \text{LLM.Rephrase}(best\_context)$;
		\State $\text{recaps.append}(\texttt{<re>} + refined + \texttt{</re>})$;
		\EndFor
		\State \Return recaps
	\end{algorithmic}
\end{algorithm}

After selecting the most relevant segment, we refine it using a LLM to enhance coherence and semantic clarity. The refined segment, termed a "recap", is enclosed with special tags $<\texttt{re}>$ and $</\texttt{re}>$ as explicit supervision signals. These recaps are further summarized into compact forms. In this process, the designed prompt emphasize that the output should be concise (typically 5 or 6 sentences) while covering all essential information.

\subsection{Recap Agent}
\begin{figure}[h!]
    \centering
    \includegraphics[width=0.8\linewidth]{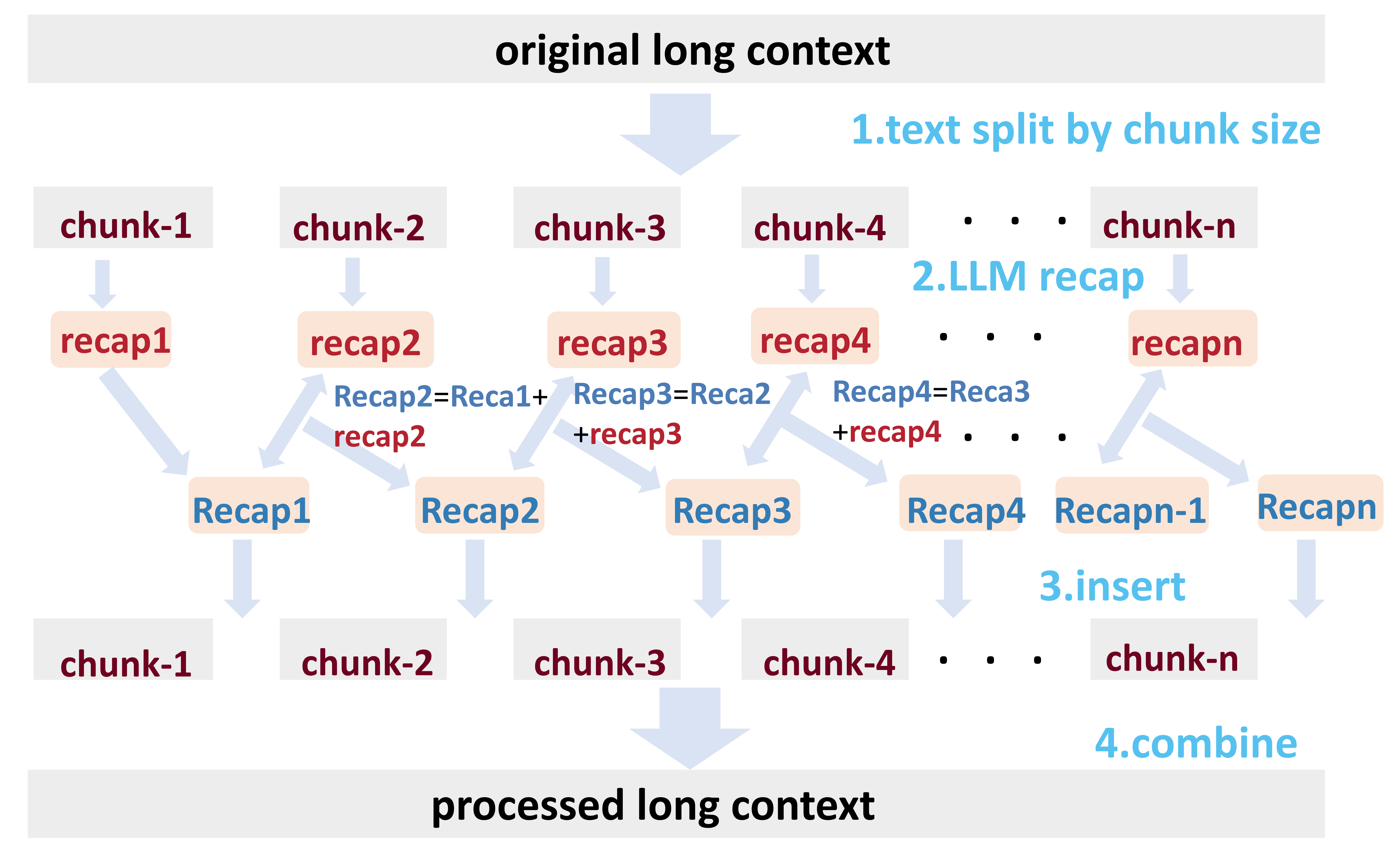}
    \caption{Full process of inference-time recap generation
and accumulation}
    \label{fig:placeholder}
\end{figure}

As shown in Fig.~\ref{fig:placeholder}. After continued pretraining on mined sequences. To leverage recap segments at inference, we introduce the recap agent. The input document is divided into non-overlapping chunks, each appended with a $<\texttt{re}>$ token to prompt the model to summarize key information.

When the accumulated recap contents exceeds the threshold $N$, the earliest portions are summarized to prevent uncontrolled context growth while retaining salient historical information. This dynamic truncation ensures effective long-range reasoning and memory retention.

\section{Experiments}
\subsection{Experimental Setups}
We evaluate the proposed ARL method on two widely used benchmarks: RULER \cite{hsieh-8} (which includes NIAH and QA tasks, among others) and LongBench \cite{bai-9} (a multi-domain long-context evaluation suite). For both benchmarks, the maximum score for each subtask is set to 100 points. On the RULER, we evaluate sequences of 8K, 16K, and 32K tokens to assess performance under different context lengths, using two representative models: Qwen 1.5 (1.8B) \cite{bai-10} and RWKV7 (1.5B) \cite{peng-11}. Prior to evaluation, we apply sample mining and perform continued pretraining of ARL on the ProLong dataset.

For pretraining, we mine long-context samples (each exceeding 10K tokens) and conduct continued pretraining on 4 $\times$ NVIDIA A800 80GB GPUs, with a learning rate of 1e-5 for 1 epoch and a batch size of 8. To accelerate training, we adopt DeepSpeed and BF16 mixed precision. During evaluation, we employ vLLM to enable efficient parallelization and maximize GPU utilization.

\subsection{Main Results}
\begin{figure}[t]
	\centering
	\includegraphics[width=\linewidth]{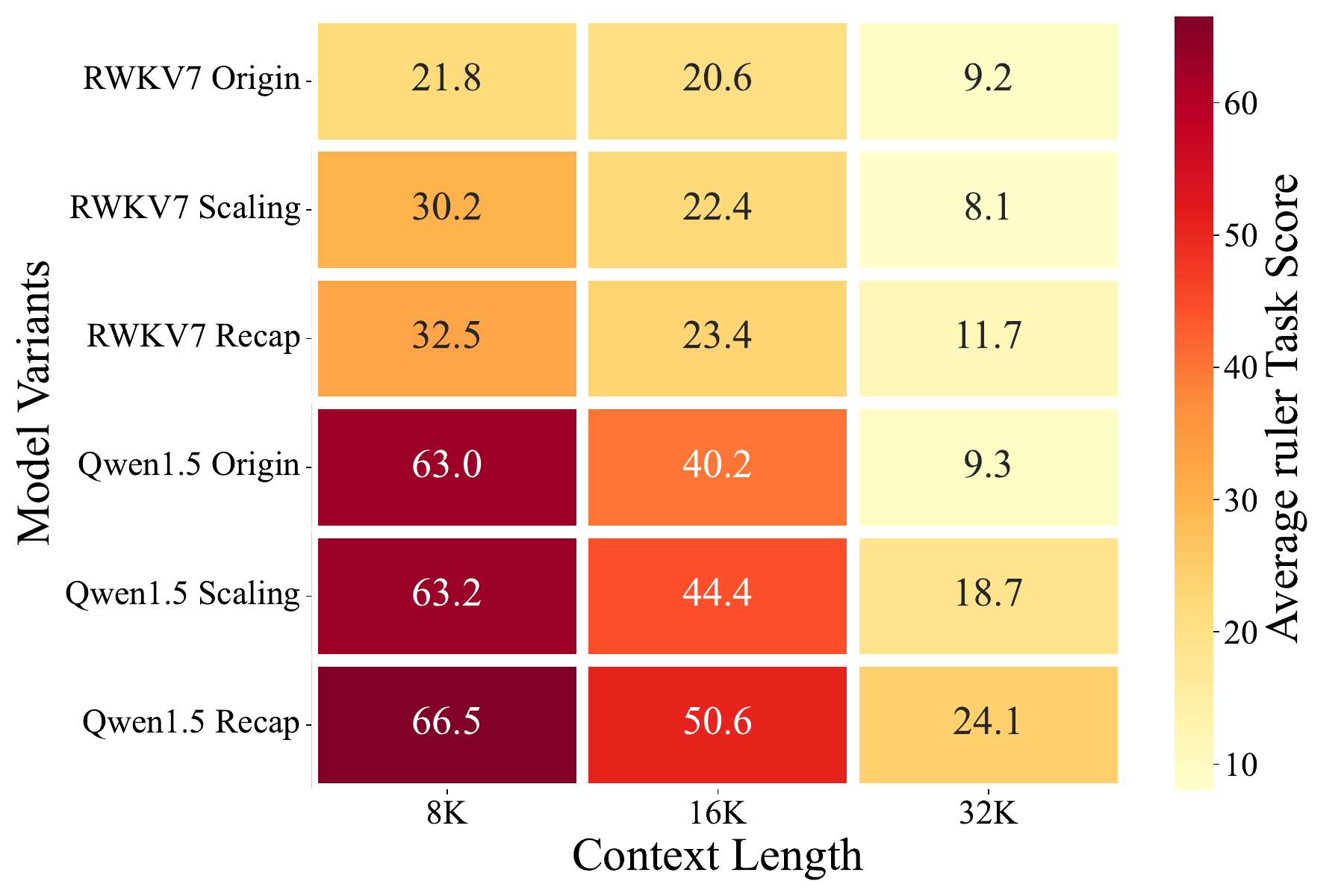}
	\caption{Performance comparison of our proposed recap with origin and scaling on RULER bench.}
	\label{figure2}
\end{figure}

As illustrated in Fig.~\ref{figure2}, ARL-trained model (Recap) consistently outperforms both the origin model and the enhanced method with effective long-context scaling, i.e., continued pretraining on the same long documents without recap augmentation. The relative improvements are highlighted in Table, where ARL yields up to 56.5\% and 129.7\% gains for RWKV7 \cite{peng-11} and Qwen1.5 \cite{bai-10}, respectively (21.4 to 33.5 at 8K and 10.1 to 23.2 at 32K). Detailed subtask-level results in Table~\ref{table2} further confirm these trends, and the recap-trained models improve most on subtasks that exploit information spread across the input, while saturated single-needle tasks leave little headroom.

On LongBench \cite{bai-9}, ARL also demonstrates consistent average improvements across the 21 tasks. Qwen1.5 achieves a 12.6\% relative gain (14.3 → 16.1) and RWKV7 improves by 11.1\% (21.6 → 24.0) on average. Particularly in summarization tasks, ARL shows substantial benefits, as reported in Table~\ref{table3}, with relative improvements ranging from 19.0\% to 47.2\%. These results collectively validate the effectiveness of ARL in enhancing long-context understanding.

\begin{table*}[htbp]
	\centering
	\setlength{\tabcolsep}{4pt}
	\renewcommand{\arraystretch}{1.2}
	\resizebox{0.88 \textwidth}{!}{
		\begin{tabular}{cccccccccccccc}
			\toprule
			\textbf{Model (Setting) / Length} & \multicolumn{3}{c}{\textbf{Multikey}} & \textbf{Query} & \textbf{Value} & \multicolumn{3}{c}{\textbf{Single}} & \textbf{CWE} & \textbf{FWE} & \textbf{Hotpot} & \textbf{SQuAD} & \textbf{VT} \\
			\cmidrule(lr){2-4} \cmidrule(lr){7-9}
			& 1 & 2 & 3 & & & 1 & 2 & 3 & & & & & \\
			\midrule
			RWKV7 1.5B (recap, 8K)  & 37.2 & 4.0 & 2.0 & 16.5 & 32.9 & 100.0 & 65.8 & 38.8 & 11.1 & 46.1 & 15.6 & 19.5 & 46.1 \\
			RWKV7 1.5B (recap, 16K) & 21.8 & 0.0 & 0.0 & 8.1  & 16.7 & 100.0 & 20.8 & 19.6 & 8.3  & 36.4 & 13.2 & 15.1 & 40.9 \\
			RWKV7 1.5B (recap, 32K) & 5.8  & 0.0 & 0.0 & 2.6  & 6.7  & 23.2  & 13.0 & 8.2  & 3.3  & 32.2 & 10.8 & 13.6 & 12.1 \\
			\midrule
			RWKV7 1.5B (origin, 8K) & 10.4 & 1.0 & 0.4 & 2.5 & 1.8 & 100.0 & 17.2 & 23.4 & 0.1 & 38.1 & 20.0 & 19.1 & 43.9 \\
			RWKV7 1.5B (origin, 16K) & 8.4  & 0.0 & 0.0 & 2.7 & 2.9 & 100.0 & 10.4 & 8.8  & 0.1 & 41.9 & 19.8 & 17.4 & 40.0 \\
			RWKV7 1.5B (origin, 32K) & 3.4  & 0.0 & 0.0 & 1.3 & 2.3 & 10.8  & 4.0  & 2.2  & 0.0 & 37.9 & 15.8 & 13.8 & 9.5 \\
			\midrule
			Qwen1.5 1.8B (recap, 8K) & 87.6 & 84.4 & 40.0 & 74.0 & 75.9 & 100.0 & 100.0 & 98.0 & 31.1 & 45.3 & 28.6 & 27.8 & 57.7 \\
			Qwen1.5 1.8B (recap, 16K)& 72.2 & 47.2 & 7.6  & 29.6 & 41.7 & 99.8  & 99.8  & 94.6 & 15.0 & 43.3 & 24.8 & 23.1 & 43.7 \\
			Qwen1.5 1.8B (recap, 32K)& 22.4 & 5.2  & 1.2  & 6.8  & 13.0 & 100.0 & 34.2  &23.0 & 5.9  & 44.4 & 18.4 & 7.8  & 18.8 \\
			\midrule
			Qwen1.5 1.8B (origin, 8K) & 85.2 & 71.8 & 20.8 & 81.2 & 86.2 & 100.0 & 99.8 & 98.8 & 24.7 & 58.2 & 31.0 & 31.8 & 45.1 \\
			Qwen1.5 1.8B (origin, 16K)& 54.6 & 7.2  & 1.4  & 22.0 & 44.4 & 99.8 & 84.4 & 79.4 & 14.5 & 33.9 & 16.2 & 17.8 & 63.2 \\
			Qwen1.5 1.8B (origin, 32K)& 7.2  & 0.2  & 0.0  & 0.8  & 5.0  & 55.0 & 6.0  & 2.8  & 4.1  & 24.5 & 12.0 & 7.5  & 6.7 \\
			\bottomrule
	\end{tabular}}
	\caption{Performance comparison of different models and settings on RULER bench.}
	\label{table2}
\end{table*}

\begin{table}[t]
    \centering
    \setlength{\tabcolsep}{15pt}
    \renewcommand{\arraystretch}{1.2}
    \resizebox{0.47 \textwidth}{!}{
        \begin{tabular}{cccc}
            \toprule
            \textbf{Task} & \textbf{Origin} & \textbf{Recap} & \textbf{Improvement} \\
            \midrule
            GovReport  & 14.28 & 18.20 & $\uparrow$27.45\% \\
            QMSum  & 13.37 & 15.91 & $\uparrow$19.00\% \\
            MultiNews  & 10.13 & 14.91 & \textcolor{red!50!black}{$\uparrow$47.19\%} \\
            VCSUM  & 7.27 & 9.04 & $\uparrow$24.35\% \\
            \bottomrule
        \end{tabular}}
    \caption{Evaluation of several summarization tasks in LongBench on Qwen1.5 1.8B.}
    \label{table3}
\end{table}

\subsection{Ablation Study and Analysis}
\subsubsection{Trade-off between sample size and context length}
To assess the impact of pretraining sample size on the generalization ability of ARL, we conduct an ablation study using the Qwen1.5 \cite{bai-10} model and evaluate on RULER bench. As shown in Table~\ref{table4}, increasing the number of training samples from 10k to 40k consistently improves performance across all evaluated context lengths (8K, 16K, and 32K), with gains of 7.5, 7.5, and 3.6 points respectively (65.4 → 72.9 at 8K and 23.2 → 26.8 at 32K). These results suggest that ARL benefits from larger-scale pretraining, and that expanding the sample size yields steady and measurable gains in long-context modeling performance.

\begin{table}[t]
    \centering
    \setlength{\tabcolsep}{20pt}
    \renewcommand{\arraystretch}{1.2}
    \resizebox{0.47 \textwidth}{!}{
        \begin{tabular}{cccc}
            \toprule
            \textbf{Context Length} & \multicolumn{3}{c}{\textbf{Sample Size }} \\
            \cmidrule(lr){2-4}
            & \textbf{10k} & \textbf{20k} & \textbf{40k} \\
            \midrule
            \textbf{8k } & 65.4 & 71.2 & 72.9 \\
            \textbf{16k } & 49.4 & 52.3 & 56.9 \\
            \textbf{32k } & 23.2 & 25.1 & 26.8 \\
            \bottomrule
        \end{tabular}}
    \caption{Comparison results of different continued pretraining sample sizes and evaluation context lengths on RULER .}
    \label{table4}
\end{table}

\subsubsection{Impact of the number of chunks}
We further investigate the effect of the number of chunks on ARL’s performance by conducting experiments with the RWKV7 \cite{peng-11} model on the NIAH  single2 task at a sequence length of 16K tokens. As shown in Table~\ref{table5}, increasing the number of chunks from 3 to 10 steadily improves performance, with scores rising from 13.4 to 19.0. This trend suggests that using more chunks (equivalent to shorter chunk sizes) facilitates more effective review and summarization of long texts, thereby enhancing the model’s ability to capture fine-grained contextual information.

\begin{table}[t]
    \centering
    \setlength{\tabcolsep}{10pt}
    \renewcommand{\arraystretch}{1.2}
    \resizebox{0.47 \textwidth}{!}{
        \begin{tabular}{ccccccc}
            \toprule
            \textbf{Chunks} & \textbf{3}  & \textbf{5} & \textbf{6} & \textbf{7}  & \textbf{9} & \textbf{10} \\
            \midrule
            \textbf{ Score} & 13.4   & 15.3 & 16.0 & 16.2  & 17.5 & 19.0 \\
            \bottomrule
        \end{tabular}}
    \caption{Comparison score of the number of different chunks on NIAH single2 task.}
    \label{table5}
\end{table}

\subsubsection{Ablation study on using OPENAI API to
summarize the mined fragments}
To verify and demonstrate that using the OpenAI API to
summarize mined article snippets improves the model’s
ability to understand long documents after continued pre-training on this dataset, we designed a control experiment:
We used the RWKV 5 1.5B model directly on an unsummarized sample dataset and on a sample dataset that summarized the mined snippets using the OpenAI API. The results
of the long bench evaluation showed that the average score
of the method using the OpenAI API was 23.2, while the
average score of the method not using the API was 17.7, a
31\% increase.

\subsubsection{Ablation study on using chunk policy }
To highlight the advantages of our ARL method over simply
continuing pre-training on long texts, we conducted the following ablation experiments on Qwen 1.5 1.8B using only
long text pre-training and the full process of our method
on the ruler bench. We found that the average score improvement for the full process at 8k token length was 8.35\%,
the average score improvement for 16k token length was
7.68\%, and the average score improvement for 32k token
length was 38\%. Especially for difficult tasks like ruler\_qa\_hotpot, using our method lead to an average rise of 16\%.

\section{Limitations}

To address concerns regarding external LLM dependency and potential disruptions caused by chunk-based processing, we clarify that the use of an external model during training functions as a targeted knowledge distillation process. Furthermore, while segmentation might seem to interrupt narrative flow, ARL’s recursive propagation mechanism maintains global coherence by carrying compressed historical context across chunk boundaries, effectively establishing a "semantic bridge" that preserves long-range dependencies. 

\section{Conclusion}
In this paper, an active recap learning (ARL) framework is designed to enhance long-context understanding in LLM. ARL integrates a sequence construction pipeline with an inference mechanism that enables recursive memory utilization through chunked processing and dynamic recap generation. Extensive experiments demonstrate that ARL achieves substantial performance gains, with particularly strong improvements on long-text summarization tasks. Furthermore, its effectiveness scales with continued pretraining data size and benefits from optimized chunking strategies, while maintaining consistent advantages across both Transformer-based and RWKV architectures.

\bibliographystyle{IEEEbib}
\bibliography{refs}

\end{document}